\documentclass[nohyperref]{article}

\usepackage{microtype}
\usepackage{graphicx}
\usepackage{subfigure}
\usepackage{booktabs} 

\usepackage{hyperref}

\usepackage[accepted]{icml2022}

\usepackage{amsmath}
\usepackage{amssymb}
\usepackage{mathtools}
\usepackage{amsthm}

\usepackage{graphicx}
\usepackage{amsmath}
\usepackage{amssymb}
\usepackage{booktabs}
\usepackage{multirow}

\usepackage[capitalize,noabbrev]{cleveref}

\theoremstyle{plain}

\theoremstyle{definition}

\theoremstyle{remark}

\usepackage[textsize=tiny]{todonotes}

\icmltitlerunning{Not All Lotteries Are Made Equal}

\begin{document}

\twocolumn[
\icmltitle{Not All Lotteries Are Made Equal}



\icmlsetsymbol{equal}{*}

\begin{icmlauthorlist}

\icmlauthor{Surya Kant Sahu}{skit,tlm}
\icmlauthor{Sai Mitheran}{tlm,nitt}
\icmlauthor{Somya Suhans Mahapatra}{tlm}
\end{icmlauthorlist}

\icmlaffiliation{skit}{Skit.ai, India}
\icmlaffiliation{tlm}{The Learning Machines}
\icmlaffiliation{nitt}{National Institute of Technology, Tiruchirapalli}

\icmlcorrespondingauthor{Surya Kant Sahu}{surya.oju@pm.me}

\icmlkeywords{Lottery Ticket Hypothesis, Sparsity}

\vskip 0.3in
]



\printAffiliationsAndNotice{\icmlEqualContribution} 

\begin{abstract}
The Lottery Ticket Hypothesis (LTH) states that for a reasonably sized neural network, a sub-network within the same network yields no less performance than the dense counterpart when trained from the same initialization. This work investigates the relation between model size and the ease of finding these sparse sub-networks. We show through experiments that, surprisingly, under a finite budget, smaller models benefit more from Ticket Search (TS). 
\end{abstract}

\section{Introduction and Related Work}
\label{sec:intro}

With the increasing amount of data and access to compute, several state-of-the-art research often focuses on improving performance by scaling up neural networks. This can be detrimental to the budget, with several underlying environmental implications. Amongst several efforts to reduce model size during optimization on the training set, pruning \cite{he2017channel, anwar2017structured, lee2020layer}, quantization \cite{yang2019quantization, Cai_2017_CVPR, lin2016fixed}, and auxiliary techniques such as transfer learning \cite{tan2018survey} have enabled the utilization of less compute while obtaining performant models.

The Lottery Ticket Hypothesis (LTH) \cite{frankle2018lottery} demonstrated that it is possible to remove up to 90\% of parameters without degrading performance. According to the LTH, at least one sub-network within the original dense model can attain the same (or even better) performance as the dense model when trained with the same initial parameters. These well-performing sub-networks are called \emph{winning tickets}. Following \cite{savarese2020winning}, we define two types of winning tickets: 
\begin{itemize}
\item \textbf{Best Sparse model:} The best performing sparse model regardless of its extent of sparsity.
\item \textbf{Sparsest Matching model:} The sparsest model that is at least as performant as the dense model.
\end{itemize}

\subsection{Ticket Search (TS)}

The problem of finding these highly sparse, highly performant sub-networks is called Ticket Search. TS is usually done through multiple \emph{train-prune-rewind} stages. Frankle and Carbin \cite{frankle2018lottery} show that sub-networks found with a single stage of \emph{train-prune-rewind} are consistently worse than those found using multiple stages. This is due to the aggressive pruning of a large number of weights, of which many would not have been pruned if pruned iteratively.

Prior work \cite{lee2020layer} has shown that TS with some pruning methods fails completely on ResNets but works well for linear-mode connected models such as VGG. Lee et al. \cite{lee2020layer} claim that Layer-Adaptive Magnitude Pruning (LAMP) works well for both VGG and ResNets. In this work, we use LAMP \cite{lee2020layer} for finding winning tickets. The method we use in the paper is outlined in \cref{ticket-search}.

\begin{algorithm}[t]
\caption{Winning Ticket Search} \label{ticket-search}
\begin{algorithmic}[1]
    \STATE Initialize weights $\theta_0$
    \STATE $\theta_{init} \leftarrow \theta_0$ \COMMENT{Store initial weights}
    \FOR{$k=0$ to $K-1$}
        \STATE $\theta_k \leftarrow Train(\theta_k)$
        \STATE $M_k \leftarrow LAMP(\theta_k)$ \COMMENT{Prune}
        \STATE {$\displaystyle \theta_{k+1} \leftarrow \theta_{init} \odot \prod_{i=0}^{k} M_i $} \COMMENT{Rewind weights}
    \ENDFOR
\end{algorithmic}
\end{algorithm}

Here, $\theta$ is the set of trainable weights for a given model architecture, and $M_k[\dots] \in [0, 1]$ is a pruning mask obtained using LAMP at stage $k$.

\subsection{Trajectory Length} 

Raghu et al. \cite{Raghu2017OnTE} propose a measure for the expressive power of a neural network. A circular trajectory/arc (in 2D) is fed to a model, and the model’s output is projected to two dimensions. The trajectory length is simply the length of the arc after transformation by the model, as observed at the output. The authors note that the trajectory length increases as the model is trained and grows exponentially with depth. Throughout this paper, we use trajectory length to quantify the change in a network's expressive power through each pruning stage.

\begin{figure}[t]
    \centering
    \includegraphics[width=0.7\linewidth]{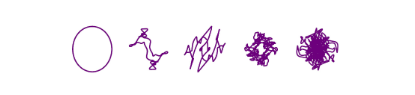}
    \caption{The 2D projections of the arc as it progresses through the neural network. Image adapted from \cite{Raghu2017OnTE}.}
    \label{traj}
\end{figure}

\paragraph{Contributions}
We present evidence through experiments on image classification (CIFAR-10) that smaller architectures consistently benefit more from TS than relatively bigger and deeper architectures. 
\section{Experimental Setup}

\label{sec:expts}

\subsection{Metrics}

In this section, we define the metrics used for our experiments.
\vspace{-1em}

  \paragraph{Success Rate ($SR$)} $SR$ denotes the \emph{difficulty} of TS for a particular model architecture. Here, $N_{success}$ is the number of times the TS yields at least one winning ticket, and $N_{total}$ is the total number of times the TS was run. 
  
  \begin{gather*}
      SR = \frac{N_{success}}{N_{total}}  \times 100
  \end{gather*}
  
  We note that $SR$ is directly influenced by $K$. Considering practical scenarios and for the purposes of this work, we set $K$ to be a small value.
  
 \paragraph{Accuracy Gain ($A_{gain}$)} $A_{gain}$ denotes how much the architecture benefits from TS. Here, $A_{sparse}$ and $A_{dense}$ are the test accuracy attained by the sparse and dense models, respectively. 
 
 Under an infinite computing budget, $A_{gain}$ denotes the \emph{value} of TS for a given architecture.

 \begin{gather*}
     A_{gain} = \frac{(A_{sparse} - A_{dense})}{A_{dense}} \times 100
 \end{gather*}

 \paragraph{Lottery TS Score ($LTS \ Score$)} $LTS \ Score \in [0, 100]$ encapsulates $A_{gain}$ and $SR$ for a given model architecture, quantifying the reliability of TS and the gain in accuracy for \emph{winning tickets}. 
 
 Under a finite computing budget, LTS Score denotes the \emph{value} of TS for a given architecture.
 
 \begin{gather*}
    LTS \: Score = A_{gain} \times SR
 \end{gather*}

\paragraph{Trajectory Length Gain ($TL_{gain}$)} $TL_{gain}$ denotes the gain in the expressive power of the neural network through the stages of pruning. $TL_{sparse}$ and $TL_{dense}$ are the trajectory lengths of the sparse and dense models, respectively, after training.
\begin{gather*}
    TL_{gain} = \frac{(TL_{sparse} - TL_{dense})}{TL_{dense}} \times 100
\end{gather*}

\begin{table}[t]
\centering
\scalebox{0.8}{
\begin{tabular}{cc}
\toprule
\textbf{Name} & \textbf{\# Params} \\ \midrule
LeNet-5  \cite{lecun1998gradient}      & 61K                \\ 
PNASNet-A  \cite{liu2018progressive}    & 0.13M              \\ 
PNASNet-B  \cite{liu2018progressive}    & 0.45M              \\ 
ResNet32    \cite{he2016deep}  & 0.46M              \\ 
MobileNetV1  \cite{howard2017mobilenets}  & 3.22M              \\ 
EfficientNet  \cite{tan2019efficientnet} & 3.59M              \\ 
ResNet18  \cite{he2016deep}      & 11.17M             \\ \bottomrule
\end{tabular}}
\caption{Chosen networks and their respective number of parameters to represent the size of the model}
\label{tab:params}
\end{table}

\subsection{Model Architectures}

We investigate six different model architectures. In Table \ref{tab:params}, we list the chosen networks and their respective number of parameters to represent the size of the model. Based on their official implementation, these networks are implemented in PyTorch \cite{paszke2019pytorch} for further experiments. Following the original ResNet paper \cite{he2016deep}, ResNet18 corresponds to the ImageNet variant of the network, whereas ResNet32 (often referred to as ResNet-S) refers to the CIFAR10 version as described in the original paper.

\section{Results}

Across all experiments, we use the default initialization scheme in PyTorch, i.e., Kaiming Uniform \cite{he2015delving}. For all models, we prune away 16\% of the weights at each pruning stage.

For the sake of repeated experiments, i.e., training many models, we set $K=5$, which is a conservative value considering that the authors of LAMP used $K=20$. This means that the TS process is not as effective since a large number of weights are pruned at each stage. However, we use the same setup for each architecture. We perform TS $50$ times for each architecture on the CIFAR-10 dataset. We compute trajectory lengths, percentage of surviving weights, test accuracy, etc., after each stage of pruning. Refer to \cref{tab:summarized} for summarized statistics for the experiments in this section.

\subsection{Ticket Search Difficulty}

With the exception of LeNet-5, most smaller architectures have above 80\% success rate. This increase in the difficulty of TS with increasing depth is consistent with results in prior work \cite{Frankle2019TheLT}.

\begin{figure}[t]
    \centering
    \includegraphics[width=\linewidth]{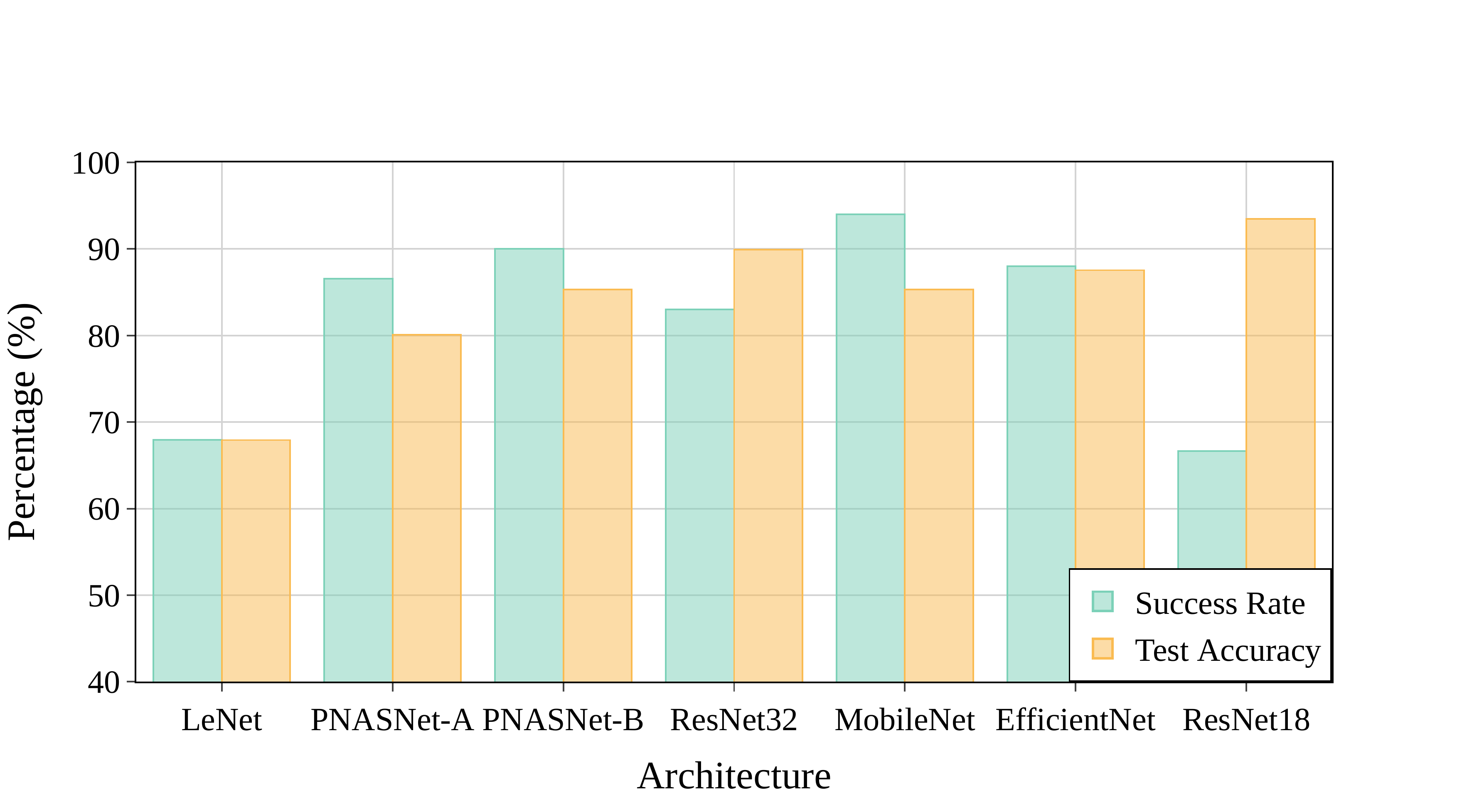}
    \caption{Ease of finding winning tickets, measured in terms of success rate. Architectures sorted in ascending order of \# parameters.}
    \label{fig:ease}
\end{figure}

\vspace{-0.5em}
For smaller models, it is possible to have a larger $K$, which would lead to an even higher success rate without a significant increase in the computing budget.

In the following subsections, we discuss the \emph{quality of winning tickets} found through TS for various architectures, building on the results of \cref{fig:ease}.

\subsection{Quality of Winning Tickets}

\paragraph{Accuracy Gain} In \cref{fig:q1}, we plot $A_{gain}$ for \emph{Best Sparse models} for runs when the TS is successful. From \cref{fig:ease} \& \cref{fig:q1}, we make the following observations:

\begin{figure}[h]
    \centering
    \includegraphics[width=\linewidth]{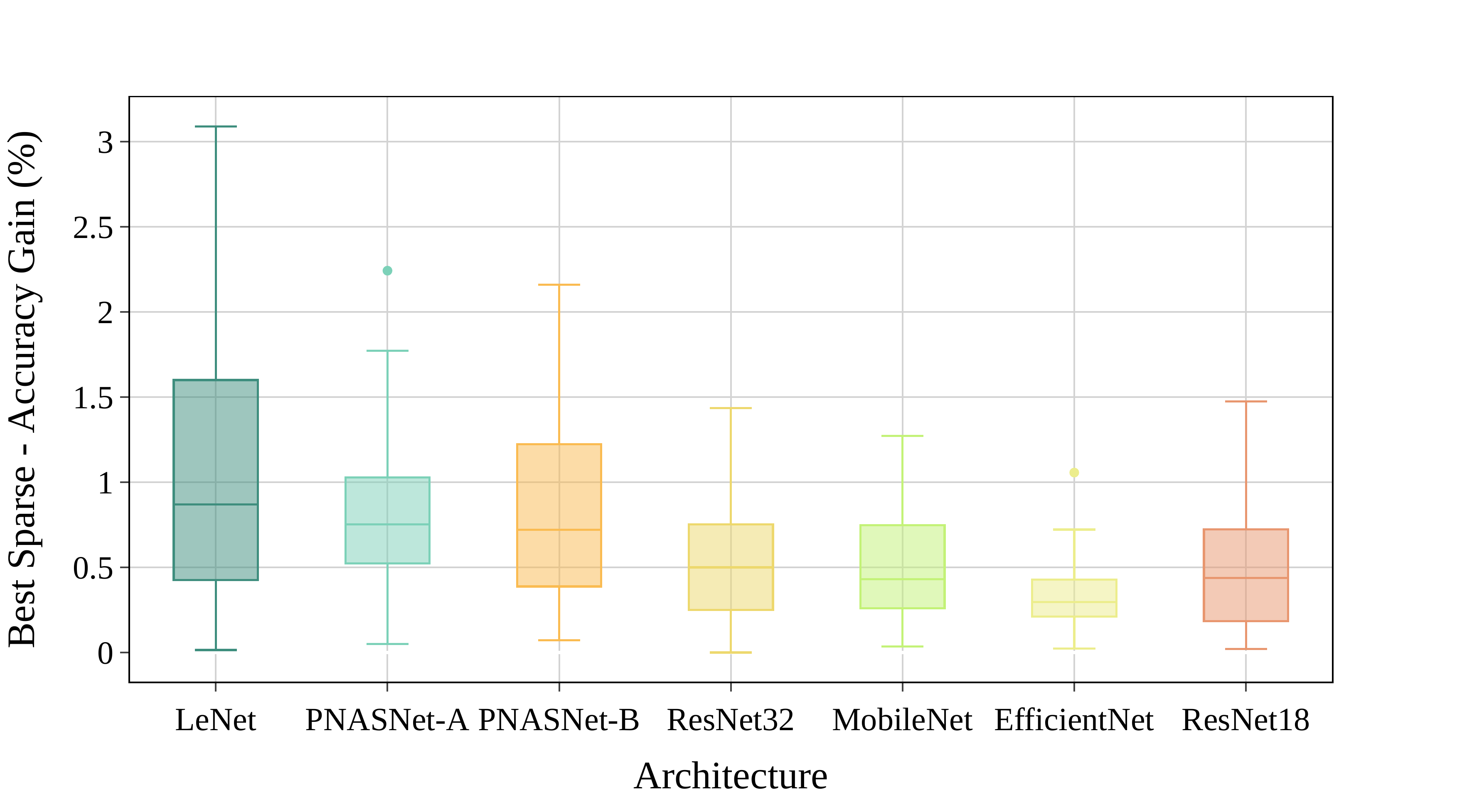}
    \caption{Accuracy gain (\%) obtained with the Best Sparse models. The smallest architecture, LeNet-5, shows the most prominent gain in accuracy, as displayed in the figure. Architectures sorted in ascending order of \# parameters.}
    \label{fig:q1}
\end{figure}

\begin{itemize}
    \item ResNet18 is the largest in terms of depth and the number of parameters. Being the largest, it has a low success rate, i.e., it is hard to find winning tickets. 
    
    \item LeNet-5 has a low success rate. However, when it does find winning tickets, the gain in accuracy is high. 
\end{itemize}

\paragraph{Which architectures benefit the most from TS?}

\begin{figure}[t]
    \centering
    \includegraphics[width=\linewidth]{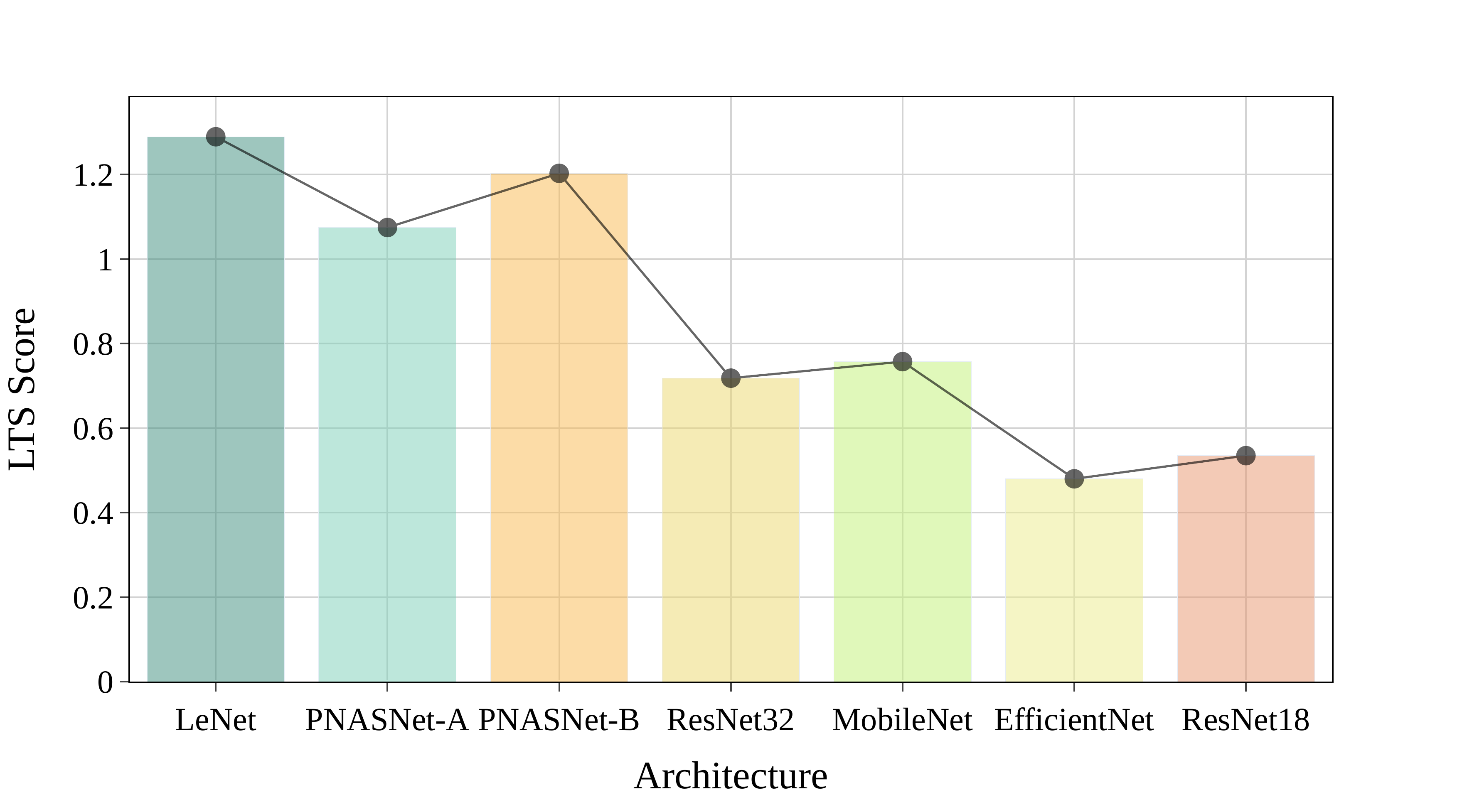}
    \caption{The decreasing trend of LTS Score with an increase in model size. From the above figure, we evidence that smaller models benefit from TS much more than larger models. Architectures sorted in ascending order of \# parameters.}
    \label{fig:q4}
\end{figure}

We define a metric, LTS Score, to encapsulate how much an architecture benefits from TS. In \cref{fig:q4}, we plot the LTS Score of each architecture. We observe a decreasing trend of LTS Score w.r.t. number of parameters.

In other words, \emph{TS can be used effectively for very small model architectures as well!}

\begin{figure}[h]
    \centering
    \includegraphics[width=\linewidth]{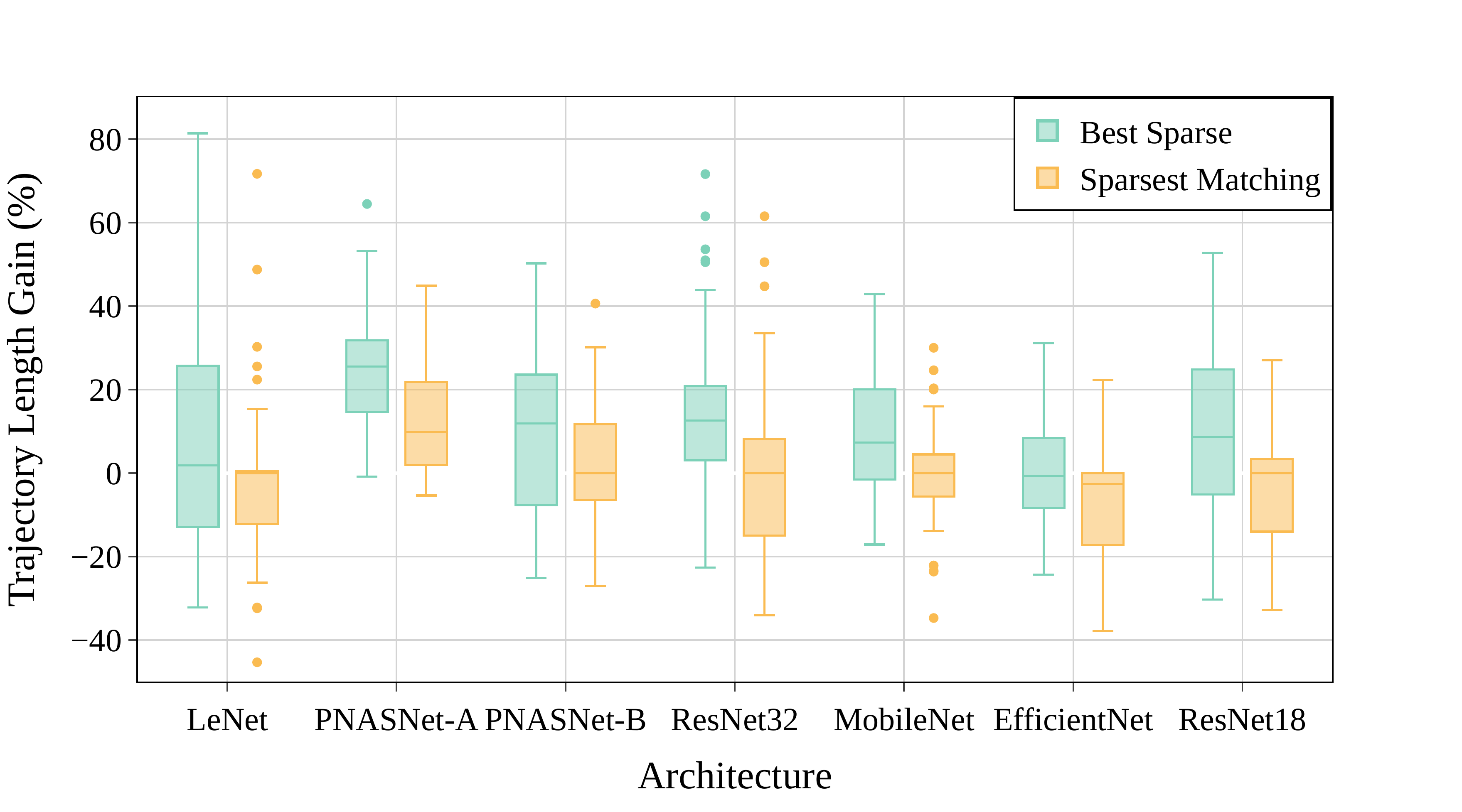}
    \caption{Trajectory length gain (\%) for Best Sparse and Sparsest Matching models, accounting for the expressivity of the models. Architectures sorted in ascending order of \# parameters.}
    \label{fig:q2}
\end{figure}

\paragraph{Expressive Power of Winning Tickets} In \cref{fig:q2}, we plot $TL_{gain}$ for both types of winning tickets, i.e., Best Sparse and Sparsest Matching models. We observe that for all architectures, the expressive power \emph{increases} for the Best Sparse models. On the contrary, the expressive power \emph{decreases} for Sparsest Matching models.

\begin{figure}[h]
    \centering
    \includegraphics[width=\linewidth]{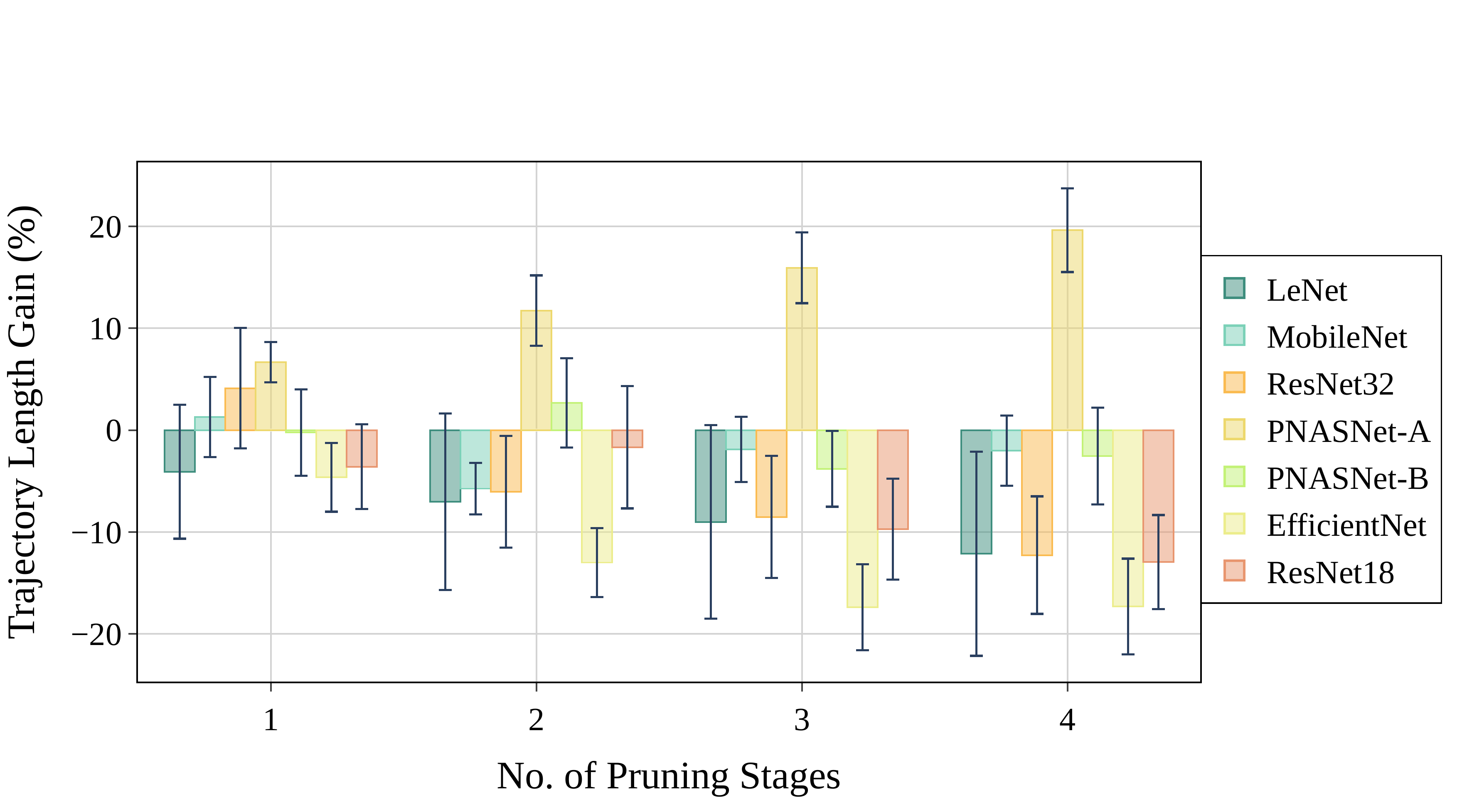}
    \caption{Trajectory length gain (\%) variation with the number of pruning stages. Architectures sorted in ascending order of \# parameters.}
    \label{fig:q3}
\end{figure}

\cref{fig:q3} illustrates how $TL_{gain}$ evolves through pruning stages. For most architectures, as expected, the $TL_{gain}$ reduces after each stage, except for PNASNets.

\begin{table*}[t]
\centering
\caption{Summarized results for all chosen architectures. Architectures are sorted according to increasing number of parameters. We observe that LTS Score progressively decreases with model size. \\}
\scalebox{0.78}{
\begin{tabular}{cccccccc}
\toprule
\multirow{2}{*}{Architecture} & \multirow{2}{*}{\begin{tabular}[c]{@{}c@{}}Network\\Type\end{tabular}} & \multirow{2}{*}{\begin{tabular}[c]{@{}c@{}}Trajectory\\Length\end{tabular}} & \multirow{2}{*}{\begin{tabular}[c]{@{}c@{}}Trajectory\\Length Gain \%\end{tabular}} & \multirow{2}{*}{\begin{tabular}[c]{@{}c@{}}Test\\Accuracy\end{tabular}} & \multirow{2}{*}{\begin{tabular}[c]{@{}c@{}}Accuracy\\Gain \%\end{tabular}} & \multirow{2}{*}{\begin{tabular}[c]{@{}c@{}}Success\\ Rate\end{tabular}} & \multirow{2}{*}{\begin{tabular}[c]{@{}c@{}}\textbf{LTS Score}\end{tabular}} \\
                              &                                                                         &                                                                              &                                                                                   &                                                                          &                                                                          &                                                                        
\\
\midrule
\multirow{3}{*}{LeNet-5 \cite{lecun1998gradient}}  & Dense                 & $0.09 \pm 0.03$                   & \ $0.00 \pm 0.00$                                      & $67.92 \pm 0.96$                 & $0.00 \pm  0.00$                              & \multirow{3}{*}{68\%}  & \multirow{3}{*}{1.2895}       \\
                              & Best Sparse           & $0.09 \pm 0.03$                    & $0.003 \pm 0.02$                         & $68.65 \pm 0.89     $             & $1.08 \pm 0.81$                   &                               \\
                              & Sparsest Matching     & $0.08 \pm 0.03$                   & -$0.004 \pm 0.02  $                        & $68.40 \pm 0.89     $            & $0.72 \pm 0.91$                  &                               
\\
\midrule
\multirow{3}{*}{PNASNet-A \cite{liu2018progressive}}  & Dense                 & $264.60 \pm 26.04$                   & \ $0.00 \pm 0.00$                                      & $80.10 \pm 0.55$                 & $0.00 \pm  0.00$                              & \multirow{3}{*}{87\%}  & \multirow{3}{*}{1.0748}       \\
                              & Best Sparse           & $321.56 \pm 21.63$                    & \ $24.35 \pm 13.37$                         & $80.75 \pm 0.51     $             & $0.80 \pm 0.44$                   &                               \\
                              & Sparsest Matching     & $290.33 \pm 27.50$                   & \ $12.04 \pm 12.32  $                        & $80.55 \pm 0.54     $            & $0.56 \pm 0.46$                  & \\
\midrule
\multirow{3}{*}{PNASNet-B \cite{liu2018progressive}}  & Dense                 & $89.08 \pm 12.38$                   & \ $0.00 \pm 0.00$                                      & $84.57 \pm 0.43$                 & $0.00 \pm  0.00$                              & \multirow{3}{*}{90\%}   & \multirow{3}{*}{1.2030}       \\
                              & Best Sparse           & $97.30 \pm 10.74$                    & \ $10.75 \pm 18.71$                         & $85.27 \pm 0.32     $             & $0.69 \pm 0.43$                   &                               \\
                              & Sparsest Matching     & $90.80 \pm 13.13$                   & \ $2.48 \pm 14.50  $                        & $84.87 \pm 0.42     $            & $0.30 \pm 0.47$                  &                              
\\ \midrule
\multirow{3}{*}{ResNet32 \cite{he2016deep}}     & Dense                 & $48.16 \pm 7.75$                     & \ $0.00 \pm 0.00       $                     & $89.90 \pm 0.25   $              & $ 0.00 \pm 0.00$                              & \multirow{3}{*}{83\%}    & \multirow{3}{*}{0.7183}     \\
                              & Best Sparse           & $55.02 \pm 8.78$                      & \ $15.33 \pm 21.17$                           & $90.36 \pm 0.24       $          & $0.52 \pm 0.35$                   &                               \\
                              & Sparsest Matching     & $48.09 \pm 8.87$                     & \  \ $0.65 \pm 20.66$                          & $90.20 \pm 0.26        $         & $0.34 \pm 0.35$                  &       
                              \\ \midrule
\multirow{3}{*}{MobileNetV1 \cite{howard2017mobilenets}}  & Dense                 & $204.81 \pm 23.62$                   & \ $0.00 \pm 0.00$                                      & $85.29 \pm 0.31$                 & $0.00 \pm  0.00$                              & \multirow{3}{*}{94\%} & \multirow{3}{*}{0.7575}        \\
                              & Best Sparse           & $220.54 \pm 26.55$                    & \ \ $8.86 \pm 13.60$                         & $85.71 \pm 0.26     $             & $0.49 \pm 0.31$                   &                               \\
                              & Sparsest Matching     & $201.61 \pm 22.67$                   & \ -$0.39 \pm 12.23  $                        & $85.46 \pm 0.32     $            & $0.21 \pm 0.36$                  &   
\\
\midrule
\multirow{3}{*}{EfficientNet \cite{tan2019efficientnet}}  & Dense                 & $331.24 \pm 43.58$                   & \ $0.00 \pm 0.00$                                      & $87.54 \pm 0.23$                 & $0.00 \pm  0.00$                              & \multirow{3}{*}{88\%}   & \multirow{3}{*}{0.4801}       \\
                              & Best Sparse           & $327.18 \pm 41.58$                    & \ \ $0.95 \pm 12.59$                         & $87.85 \pm 0.22     $             & $0.35 \pm 0.2$                   &                               \\
                              & Sparsest Matching     & $303.08 \pm 50.72$                   & \ -$6.67 \pm 14.52  $                        & $87.65 \pm 0.25     $            & $0.12 \pm 0.24$                  &                                
                              \\ \midrule
\multirow{3}{*}{ResNet18 \cite{he2016deep}}     & Dense                 & $22.60 \pm 2.76$                     & \ $0.00 \pm 0.00       $                     & $93.47 \pm 0.33   $              & $ 0.00 \pm 0.00$                              & \multirow{3}{*}{67\%}   & \multirow{3}{*}{0.5350}       \\
                              & Best Sparse           & $24.43 \pm 3.22$                      & \ $10.79 \pm 20.73$                           & $93.90 \pm 0.16       $          & $0.46 \pm 0.34$                   &                               \\
                              & Sparsest Matching     & $21.48 \pm 2.67$                     & \ -$3.02 \pm 14.94$                          & $93.69 \pm 0.26        $         & $0.16 \pm 0.33$                  &    
                              \\
\bottomrule        
\end{tabular}}
\label{tab:summarized}
\end{table*}

We note that PNASNets are designed using Neural Architecture Search (NAS) \cite{zoph2016neural} with the objective of maximizing validation accuracy and are not hand-designed by humans, unlike other models experimented with in \cref{tab:params}.

The performance on the validation set is used as the early stopping criterion at each \emph{train-prune-rewind} stage. By virtue of early stopping, the Sparsest Matching models perform on-par with the Dense model on validation \& test sets. However, from \cref{tab:summarized}, $TL_{gain}$ of Sparsest Matching models is near-zero or negative for most architectures. This suggests that the TS procedure yields a model that tends to overfit on the validation set, evident by the decrease in the expressive power. In other words, TS should not be used to prune models to extremely high sparsity levels, as it might negatively impact the model's robustness.

In future work, we plan to analyze the relationship between $TL_{gain}$ and robustness to noise and distribution shifts.

\section{Conclusion}
This work investigates the relationship between model size and the ease of finding winning tickets and study their expressive power. Through experiments, we make the following observations: a) smaller models benefit more from TS, b) mild pruning increases the expressive power of the network and lowers generalization gap, c) excessive pruning reduces model expressivity while retaining test-set performance, which may lead to poor robustness. We hope that this work will encourage industry practitioners to use TS to increase the performance of real-time, lightweight models.   

\nocite{langley00}

\bibliography{egbib}
\bibliographystyle{icml2022}



\end{document}